\documentclass[journal]{IEEEtran}

\usepackage{booktabs}
\usepackage{multirow}
\usepackage{arydshln}
\usepackage{makecell}
\usepackage{hyperref}
\usepackage{float}

\usepackage{cite}

\usepackage{graphicx}

\begin{document}

\title{Multi-Record Web Page Information Extraction From News Websites}

\author{
\IEEEauthorblockN{Alexander Kustenkov$^{1,2}$, Maksim Varlamov$^{1}$, and Alexander Yatskov$^{1,2}$}\\
\IEEEauthorblockA{$^1$\textit{Ivannikov Institute for System Programming of the Russian Academy of Sciences, Moscow, Russia}}\\
\IEEEauthorblockA{$^2$\textit{Lomonosov Moscow State University, Moscow, Russia}\\
\{kustenkov, varlamov, yatskov\}@ispras.ru}
}

\maketitle

\begin{abstract}
In this paper, we focused on the problem of extracting information from web pages containing many records, a task of growing importance in the era of massive web data. Recently, the development of neural network methods has improved the quality of information extraction from web pages. Nevertheless, most of the research and datasets are aimed at studying detailed pages. This has left multi-record "list pages" relatively understudied, despite their widespread presence and practical significance.

To address this gap, we created a large-scale, open-access dataset specifically designed for list pages. This is the first dataset for this task in the Russian language. Our dataset contains 13,120 web pages with news lists, significantly exceeding existing datasets in both scale and complexity. Our dataset contains attributes of various types, including optional and multi-valued, providing a realistic representation of real-world list pages. These features make our dataset a valuable resource for studying information extraction from pages containing many records.

Furthermore, we proposed our own multi-stage information extraction methods. In this work, we explore and demonstrate several strategies for applying MarkupLM to the specific challenges of multi-record web pages. Our experiments validate the advantages of our methods.

By releasing our dataset to the public\footnote{https://github.com/ispras/news-page-dataset}, we aim to advance the field of information extraction from multi-record pages.
\end{abstract}

\begin{IEEEkeywords}
Data collection, Data extraction, Dataset, Multi-record extraction, Neural network
\end{IEEEkeywords}

\IEEEpeerreviewmaketitle

\def\ItIsIEEEDoc{1} 
\section{Introduction}
\ifx\ItIsIEEEDoc\undefined In \else In \fi the digital age, a huge amount of information presented on the Internet often exists in complex multi-record formats, especially on news websites. A record is defined as a separate entity on a webpage, for example, for news resources it is one news item, that includes a title, publication date and possibly other characteristics. A multi-record page is a page containing several records. In the news domain, this might be a page of some category, which usually displays several news items. A page with only a single record (news item) is called detailed. Extracting data from multi-record pages poses a significant problem, especially given that traditional methods and datasets are mostly focused on single-record or detailed pages. This paper presents a novel approach to extracting information from multi-record web pages, especially those located on Russian-language news websites. This article aims to improve the accuracy and efficiency of data extraction from such intricate web environments by building a large dataset of over 13,000 pages and applying neural methods such as the MarkupLM model. Despite the numerous approaches to solving the problem of information extraction, this article will focus on methods that do not require a visual representation of the page (including generating a representation from the HTML code) and instead rely only on HTML code.

\section{Related work}

There are several datasets for the task of extracting attributes from web pages with many records. They all have their own unique features, since each was collected to test a specific method of solving this problem. However, there is a certain set of characteristics by which we can compare them:
\begin{itemize}
    \item Number of pages for each subject area
    \item Number of sites for each subject area
    \item Number of records for each subject area
    \item Number of pages in total
    \item Year of dataset release
    \item Language of dataset
\end{itemize}

One of the first datasets was announced in the article ``Simultaneous Record Detection and Attribute Labeling in Web Data Extraction``\cite{zhu_simultaneous_2006} in 2006 and was named LDST -- ListDataSeT. In this dataset, the authors included 771 pages which contained lists of records.

The first large dataset aimed at studying the problem of extracting information from pages with many records was proposed in the paper ``AMBER: Automatic Supervision for Multi-Attribute Extraction``\cite{furche_amber_2012} in 2012. The authors of the article focused on 150 real sites that contained pages with many records. The resulting dataset contained more than 2,000 pages with a total number of records more than 20,000.

In 2020, a team of researchers from Amazon published the article ``PLAtE: A Large-scale Dataset for List Page Web Extraction``\cite{san_plate_2023} in which they presented a new dataset PLATE -- Pages of Lists Attribute Extraction which contains pages of online stores. The authors focused on preparing high-quality data, they conducted a multi-stage preparation of data, which included:
\begin{enumerate}
 \item Filtration of pages that did not contain lists of records.
 \item Among other pages, the most ``popular`` websites were selected based on data published in the ``Tranco List``.
 \item Also, preference was given to sites that contained as many pages with lists as possible.
 \item Pages with obscene content were filtered.
\end{enumerate}
Thus, the dataset was based on 43 sites and more than 6,600 pages, which together contains more than 52,000 records. Based on this, main feature of this dataset is the high quality of the data.

One of the problems related to web page extraction is web page segmentation. This task involves finding the boundaries of each record on the page. There are several approaches to the segmentation problem. Some of them assume the presence of visual information. For example, such algorithms are discussed in the article \cite{kiesel2021empirical}. However, in this work, we assume that visual information is unavailable and we will use other methods.

One of the first solutions of finding record boundaries on the page is the algorithm proposed in the article ``Web data extraction based on partial tree alignment``\cite{zhai_web_2005}. The authors developed algorithm called MDR, this algorithm finds ``similar`` fragments on the page. It is assumed that "similar" fragments records have the same attributes. After finding these fragments, the process of ``alignment`` occurs, this process compares each node of the fragment to the node in the remaining fragments. By this algorithm we can present information in a structured form.

It is also worth mentioning the article ``Finding and Extracting Data Records from Web Pages``\cite{alvarez_finding_2010}, in which an algorithm for generating partition candidates and selecting the best one was proposed. In our article we consider using this algorithm from a practical point of view.

\begin{figure}
    \centering
    \includegraphics[width=0.45\textwidth]{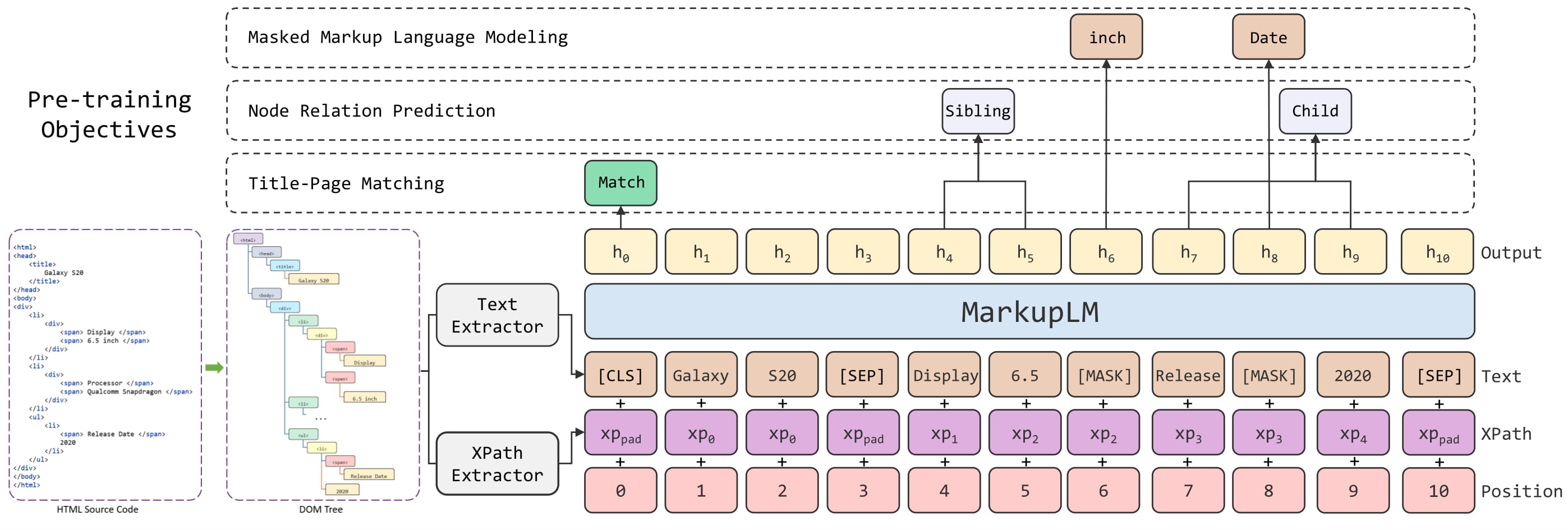}
    \caption{MarkupLM architecture*}
    \small* This picture is taken from the \cite{li_markuplm_2022}.
    \label{fig:markuplm_architecture}
\end{figure}

According to our research, there are information extraction methods designed to extract from detailed pages, which might be generalized for the task of extracting information from list pages. One of these methods is the MarkupLM, which was designed in 2022 by Microsoft\cite{li_markuplm_2022}. MarkupLM is BERT-like model, which is used for processing markup-language-based documents such as HTML or XML. The main feature of this model is encoding text and location of node together. The authors propose to use this model as a tool for working with detailed pages. More detailed description of architecture of this model is given in the \autoref{fig:markuplm_architecture}.

\begin{table*}[!ht]
\caption{Dataset comparison}
\begin{center}
    \begin{tabular}{ccccccc}
        \toprule
        Dataset & Pages/Vert. & Sites/Vert. & Records/Vert & Total Pages & Year & Language\\
        \midrule
        LDST & 771 & --- & 8600 & 771 & 2011 & English\\
        AMBER & 200 & 150 & 2000 & 431 & 2012 & English\\
        PLATE & 6694 & 43 & 52900 & 6694 & 2020 & English\\
        \midrule
        Our dataset & \textbf{13120} & \textbf{278} & \textbf{257595} & \textbf{13120} & 2023 & Russian\\
        \bottomrule
    \end{tabular}
    \label{tab:dataset_compare}
\end{center}
\end{table*}
\section{Problem definition}
Given a set $W$ of n websites. Each website $W_i$, where $1 \leq i \leq$ n, is represented by a certain number of pages $m_i$, a website may consist of just one page, i.e. $m_i = 1$. 
Each page $W_i^j$ on website $W_i$, where $j$ is the page number, with $1 \leq j \leq m_i$, is a multi-record page, meaning it contains $k$ records where $2 \leq k$. 
We defined a record as an entity with a predefined set of characteristics, e.g. in the news domain these characteristics might include date, author, title, tag, etc. So the set of record's characteristics is a vector $H = \{h_0, h_1, ..., h_t\}$, where $t$ the number of characteristics in a given record. It is important to note that a record can have multiple characteristics of the same type, such as multiple tags.

We formulate the task of extracting information from multi-record pages as identifying all vectors $H$ within the given set $W$, with the following constraints:
\begin{enumerate}
  \item The proposed methods must not rely on visual information based on page rendering.
  \item The proposed methods must be applicable to all multi-record pages in the domain, even if the website was not included in the training dataset.
  \item The proposed methods should generalize beyond the news domain to other fields.
\end{enumerate}

The output of each method should be a structured representation of the records for each page, for example in JSON format.

\section{Dataset construction}
We decided to collect our own dataset for the task of extraction information from multi-record web-pages and make it publicly available. In this chapter we describe how the data was collected and annotated, also we provide the final dataset's main characteristics.

\subsection{Data collecting}
Our dataset contains news web pages collected from Russian-language media. News resources were selected according to the MediaMetrics\footnote{https://mediametrics.ru/rating/ru/online.html} latest news quotations system, which provides rankings based on the popularity of news resources. Web pages were downloaded between 12/28/2023 and 04/28/2024. Pages were downloaded using a special Python3 script (based on the Scrapy library\cite{scrapy_library}), which was run daily through prepared sitemaps.

\subsection{Sitemaps development}
Sitemaps were prepared using the WebScraper\footnote{https://github.com/ispras/web-scraper-chrome-extension} browser extension for Chrome. Using sitemaps in special web crawlers allows to download an html page, as well as an answer for this page. Thus, each site requires the development of a unique sitemap. On each page, the boundaries of each record and its attributes were annotated, if they existed. Only the following attributes were noted: \textit{date}, \textit{title}, \textit{tag}, \textit{short\_title}, \textit{author}, \textit{time}. The annotation was done manually.
We chose this set of attributes because they are the most popular in the news websites domains. 
For each site several categories were annotated, for example, politics, economics, and sports. In this way 312 sitemaps were prepared.

\subsection{Dataset preparing}
\begin{figure}[H]
    \centering
    \includegraphics[width=0.45\textwidth]{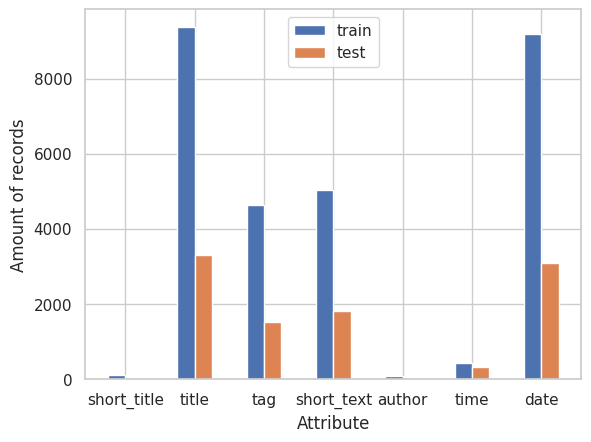}
    \caption{Attribute distribution between train and test}
    \label{fig:attribute_dist}
\end{figure}

\begin{figure}[H]
    \centering
    \includegraphics[width=0.45\textwidth]{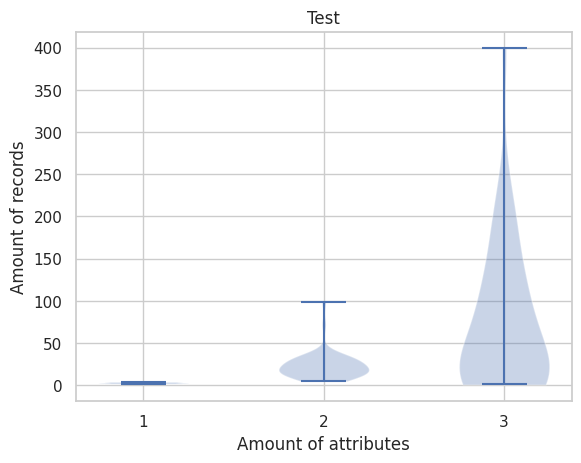}
    \caption{Distribution of amount of attributes at test part}
    \label{fig:amount_of_attributes_test}
\end{figure}

\begin{figure}[H]
    \centering
    \includegraphics[width=0.45\textwidth]{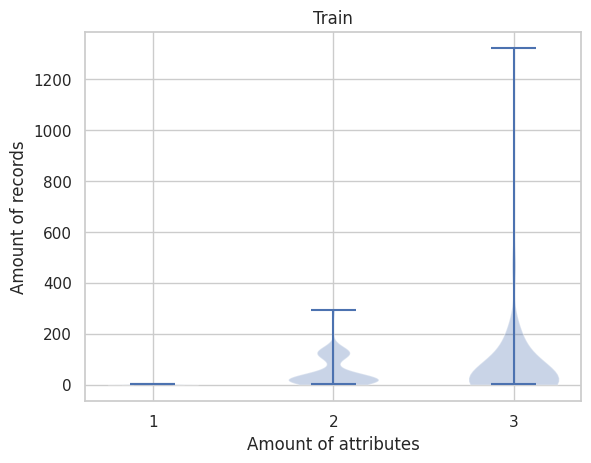}
    \caption{Distribution of amount of attributes at train part}
    \label{fig:amount_of_attributes_train}
\end{figure}

For further use of data in our dataset, it was necessary to preprocess the raw data. The following actions were carried out:
\begin{enumerate}
    \item Filtering duplicate pages. Some downloaded pages contained many records scraped before on other pages. Such pages, with more than a quarter repeated records, were filtered.
    \item Cleaning HTML. At this stage blocks, such as blocks of code in JavaScript (i.e. all nodes with the \verb|<script>| tag), that do not affect the algorithm's performance were removed from the HTML code of the page.
    \item Translating HTML. Since our dataset contains pages in Russian, it was necessary to translate them into English to be able to use pre-trained models.
    \item Division into training and test parts. The distribution of attributes and domains of web pages was taken into account. Each domain was placed either to the training or to the test parts (see attributes split in the \autoref{fig:attribute_dist}). The final distribution is shown in the \autoref{fig:amount_of_attributes_test} and \autoref{fig:amount_of_attributes_train}. Ratio of parts after splitting was the following: 75\% - training, 25\% - testing.
\end{enumerate}

\subsection{Dataset Statistics}
Since the maps were based on CSS selectors, there was a problem with downloading sites that dynamically change the names of the styles on the pages. In other words, when the website changed the name of the CSS style class, the selector specifying the class name stopped working correctly. Therefore, we were able to download only 278 sites.

\begin{figure}[H]
    \centering
    \includegraphics[width=0.45\textwidth]{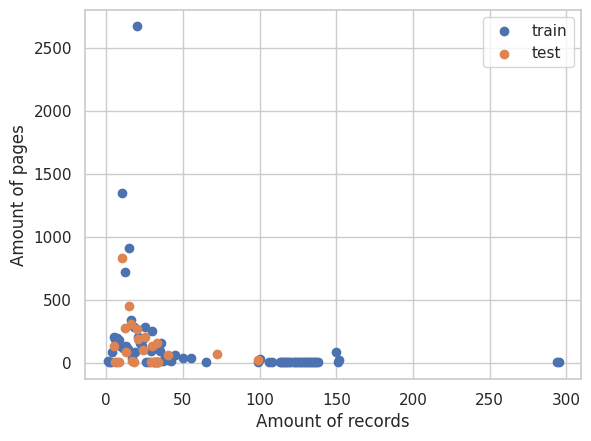}
    \caption{Distribution between pages and amount of records at page at train and test parts}
    \label{fig:page_records_dist}
\end{figure}

\begin{table}[!htb]
    \caption{Attribute frequency}
    \centering
    \begin{tabular}{c|c|c|c}
        \toprule
        Name & Pages & Records & Sites\\
        \midrule
        title & 12679 & 247262 & 275\\
        date & 12296 & 241634 & 251\\
        tag & 6165 & 108400 & 140\\
        \cdashline{1-4}
        short\_text & 6855 & 115983 & 138\\
        short\_title & 105 & 1289 & 4\\
        author & 87 & 957 & 1\\
        time & 730 & 15809 & 8 \\
        \bottomrule
    \end{tabular}
    \label{tab:attribute_statistic}
\end{table}
Thus, a dataset that contained 13120 pages from 278 Internet media was prepared (distribution between pages and entries on them, shown in the \autoref{fig:page_records_dist}). On each page, the corresponding attributes were annotated,  and their frequency was presented in the \autoref{tab:attribute_statistic}. All pages presented in the table have UTF-8 encoding. 

We researched extraction of 3 attributes from our dataset: title, author and tag. Because they are the most frequent in our dataset.
\section{Experimental evaluation}
In this chapter we describe our methods of solving the problem and show the results on our dataset.
\subsection{Parallel pipeline}
\begin{figure}[!htb]
    \centering
    \includegraphics[width=0.45\textwidth]{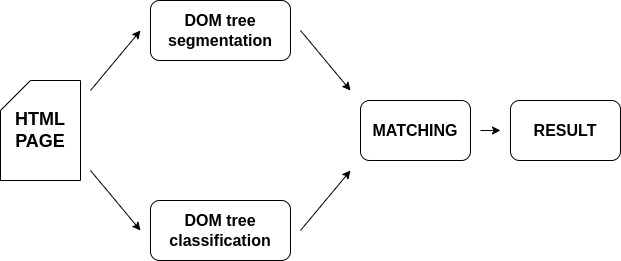}
    \caption{Parallel pipeline scheme}
    \label{fig:parallel_pipeline}
\end{figure}

The first architecture we tested was ``Parallel pipeline``. Visual scheme of this architecture described in \autoref{fig:parallel_pipeline}.
\begin{itemize}
    \item \textbf{Segmentation} At this stage, we find records boundaries that help us split html into records and find corresponding attributes of each record.
    \item \textbf{Classification} At this stage, we give a corresponding label for each node on html. It is important that the segmentation stage and the classification stage are two independent stages.The results of neither are not shared with the other.
    \item \textbf{Matching of results} At this stage, the final result of the method is formed. Based on the results of segmentation and classification, the records are matched with their attributes and provided in a structured view.
\end{itemize}

\subsection{Sequential pipeline}
\begin{figure}[!htb]
    \centering
    \includegraphics[width=0.45\textwidth]{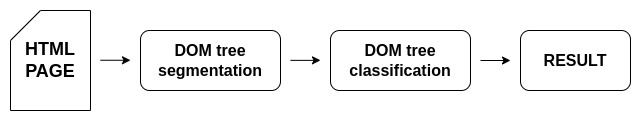}
    \caption{Sequential pipeline scheme}
    \label{fig:sequential_pipeline}
\end{figure}

We will also test sequential pipeline as described at \autoref{fig:sequential_pipeline}. In the following we will compare the qualities of both architectures. Quality should vary due to different approaches to the classification stage.
\begin{itemize}
    \item \textbf{Segmentation} In sequential architecture this stage is similar to parallel scheme.
    \item \textbf{Classification} At this stage, the final result of the method is formed. Based on the segmentation results, the stage of searching for attributes only in the selected area is carried out. Thus, the matching stage is not required.
\end{itemize}

\subsection{Segmentation subtask} \label{segmentation}
We formulate the task of web-pages segmentation as a task of searching for information boundaries, which are first nodes containing text of the DOM subtree and belonging to it. This formulation is the same as that used in the \cite{san_plate_2023}.

We test two methods of solving this task:
\begin{enumerate}
    \item Heuristic method based on classical MDR
    \item Neural method based on MarkupLM, which is state-of-the-art model in information extraction from HTML pages 
\end{enumerate}

\textbf{MDR} We tested MDR because it has open-source code. Since this method proposes several ``candidates`` for segmentation, ordered by their probability. We choose the segmentation with the highest probability.

\textbf{MarkupLM} We train the MarkupLM model on this task, having the data previously prepared: for each record on the page, we have marked its first node which contains text with the “BEGIN” label. All other nodes on the page have been marked with the label “OUT“. Thus, the solution of the problem is to predict the corresponding label for each DOM tree node by the model.

\subsubsection*{Segmentation metrics}
The result of the segmentation method can be evaluated by page-weighted metrics: $Precision_{avg}$, $Recall_{avg}$, $F1_{avg}$

To calculate them, the reference segmentation of the given page on the record is compared with the one obtained by the proposed methods. Based on this comparison, for each segment it is possible to calculate $TP_{page K}$ -- the number of DOM nodes correctly marked as an information boundary, $FP_{page K}$ - the number of DOM tree nodes that are not an information boundary, but marked with it, $FN_{page K}$ -- the number of DOM tree nodes that are an information boundary, but not marked as it:

$$
Precision_{page K} = \frac{TP_{page K}}{TP_{page K} + FP_{page K}}, 
$$

$$
Recall_{page K} = \frac{TP_{page K}}{TP_{page K} + FN_{page K}},
$$

$$
F\textit{1}_{page K} = 2 \cdot \frac{Precision_{page K} \cdot Recall_{page K}}{Precision_{page K} + Recall_{page K}},
$$

$$
Precision_{avg} = \frac{\sum_{i = 1}^{K}Precision_{page\,i}}{K},
$$

$$
Recall_{avg} = \frac{\sum_{i = 1}^{K}Recall_{page\,i}}{K},
$$

$$
F\textit{1}_{avg} = \frac{\sum_{i = 1}^{K}F\textit{1}_{page\,i}}{K}
$$

Also we will calculate classical $NMI$\cite{Strehl2002} and $ARI$\cite{Hubert1985} for segmentation task.

\begin{table}[!htb]
    \caption{Results of segmentation experiment}
    \centering
        \begin{tabular}{c|ccc|cc}
        \toprule
        Method & $Recall_{avg}$ & $Precision_{avg}$ & $F\textit{1}_{avg}$ & $ARI$ & $NMI$\\
        \midrule
         MDR & 0.473 & 0.486 & 0.465 & 0.437 & 0.517\\
         MarkupLM & \textbf{0.908} & \textbf{0.941} & \textbf{0.925} & \textbf{0.802} & \textbf{0.869}\\
         \bottomrule
    \end{tabular}
    \label{tab:segmentation}
\end{table}

The test results are shown in the \autoref{tab:segmentation}. The MarkupLM model shows superior results in all metrics in comparison with the other tool.

\subsection{Classification subtask}
This subtask includes searching for all potential attributes in the area of interest on the html page (for parallel pipeline - it is the whole page, and for sequential - a part of it).

\begin{table}[htbp]
\centering
\caption{Results of classification experiment}
\begin{tabular}{c|c|*3c}
    \toprule
        Type                                    &     Metric            & title    & tag      &  date \\
        
        \midrule
\multirow{3}{*}{Record context}                & $Precision_{avg}$     & 0.98     & 0.79     & 0.86  \\
                                                & $Recall_{avg}$        & 0.99     & 0.85     & 0.91  \\
                                                & $F\textit{1}_{avg}$   & \textbf{0.99}     & 0.82      & \textbf{0.88}  \\
        
        \midrule
\multirow{3}{*}{Page context}                   & $Precision_{avg}$     & 0.68     & 0.61       & 0.64      \\
                                                & $Recall_{avg}$        & 0.95     & 0.76       & 0.86      \\
                                                & $F\textit{1}_{avg}$   &  0.79    & \textbf{0.89}       & 0.73      \\
    \bottomrule
\end{tabular}
\label{tab:markuplm_classification}
\end{table}

The MarkupLM model is also chosen as a classification method. We train it on the task of predicting a node’s label. Thus, the model predicts the labels: ``title``, ``tag``, ``date``.

\subsubsection*{Classification metrics}
Results are evaluated by classical page-weighted classification metrics: $Precision_{avg}$, $Recall_{avg}$ and $F\textit{1}_{avg}$.

We trained two models: one for the pipeline with full page context (in such conditions, classification is used in the parallel pipeline), and one with just record context (passing the part of the html related to a specific record directly to the input, so it is used in a sequential pipeline). The comparison of their results is presented in the table \autoref{tab:markuplm_classification}. So, MarkupLM with record context shows rather high results in this subtask.

\subsection{Matching subtask}
We propose matching algorithm:
\begin{itemize}
    \item Let $G$ -- set of xpaths\footnote{We are considering positional xpath expressions consisting of tags and indices in the DOM tree path from root to the given node.} to all DOM-nodes of htmls marked as ``information boundary`` at the segmentation stage.
    \item Let's make $\widehat{G}$: for each xpath from $G$ find minimal (by length) node prefix. Each prefix should belong to only one xpath from $G$.
    \item Let's enumerate $\widehat{G}$.
    \item  Let's refer Xpath of some attribute determined at classification stage as $xpath_{attr}$. Each attribute is matched with record with number $N$ if $\widehat{G}[N]$ is prefix for $xpath_{attr}$.
\end{itemize}

\subsection{Final method evaluation metrics}
We evaluated the final method using metrics similar to detailed page information extraction task's metrics. We define ``predicted record`` as a set of predicted attributes matched to the same record at the Matching stage. Also we define ``reference record`` as a set of ground truth attributes values from a sought ``information boundary``. There are three cases possible when the method is run:
\begin{enumerate}
    \item \textit{We can match the predicted record with the reference record.}\\ $TP$ is the number of correctly extracted attributes (for example if a record contains 4 tags and they are extracted correctly, then $TP$ is 4). $FP$ is the number of extracted attributes but none of them should have been. $FN$ is the number of not extracted attributes but all of them should have been.
    \item \textit{We cannot match any of the predicted records with the reference record.}\\ Only $FN$ increases for all attributes of the reference record.
    \item \textit{We cannot match any of the reference records with the predicted record.}\\ In this case $FP$ increases for all attributes of the predicted record.
\end{enumerate}

\subsection{Experiments results}
\begin{table}
\centering
\caption{Results of final method}
\begin{tabular}{c|c|*3c}
    \toprule
        Method                  &     Metric     & title    & tag       &  date        \\
        
        \midrule
\multirow{6}{*}{Parallel Pipeline}             & TP & 56008    & 29202     & 60570        \\
                                                & FP & 7021     & 10804     & 12728       \\
                                                & FN & 8083     & 4484      & 12361       \\
                                                \cmidrule(lr){2-5}
                                                & Precision & \textbf{0.874}     & \textbf{0.867}     & \textbf{0.831}  \\
                                                & Recall    & 0.889     & 0.73      & 0.826  \\
                                                & F1        & 0.881     & 0.793     & 0.828  \\
        
        \midrule
\multirow{6}{*}{Sequential Pipeline}            & TP & 55891    & 28643     & 58645           \\
                                                & FP & 1492     & 7641      & 7056            \\
                                                & FN & 8801     & 4899      & 13659           \\
                                                \cmidrule(lr){2-5}
                                                & Precision & 0.864     & 0.854     & 0.811  \\
                                                & Recall    & \textbf{0.974}     & \textbf{0.789}     & \textbf{0.893}  \\
                                                & F1        & \textbf{0.916}     & \textbf{0.82}      & \textbf{0.85}   \\
    \bottomrule
\end{tabular}
\label{tab:final_experiments_results}
\end{table}

We tested both proposed methods: parallel pipeline and sequential pipeline. For these methods, we used the corresponding trained versions of MarkupLM, as we described them in the previous chapters. Comparative results are shown in \autoref{tab:final_experiments_results}, while \autoref{tab:markuplm_classification} presents the performance of classification model in different contexts—record and page.

In \autoref{tab:markuplm_classification}, the model using the ``record context`` outperforms model using the ``page context``. The tag attribute in the page context performs relatively well, but the overall trend shows that the model is more effective in the record context. We assume that the advantage of page context for tag-attribute extraction is related to the peculiarity of this attribute. Most often, each record has several tags. Information about neighboring records allows making less noisy predictions.

The parallel pipeline demonstrates solid precision, particularly for title. However, it shows a decline in recall. In contrast, the sequential pipeline outperforms the parallel pipeline in recall. The sequential pipeline achieves a balanced performance, making it the preferred method due to its robustness and consistency.

In conclusion, the sequential pipeline is superior to the parallel pipeline. The record context further enhances classification performance, making it the optimal setting for method deployment. The high results of the obtained methods prove their applicability in real life.

\section{Conclusion}

In this paper, we propose a new large-scale dataset consisting of over 13,000 web pages from Russian news sources, which was specifically designed to address the problem of extracting information from multi-record web pages. This dataset is the first dataset in the Russian language for information extraction from multi-record pages task. It is significantly larger than existing datasets, and includes pages with optional and multi-valued atributes. We make it openly available to provide a valuable resource for researchers of extraction methods.

We have applied and validated multi-stage methods for information extraction. In this methods we fine-tuned a state-of-the-art MarkupLM, which was designed to understand and process markup-language-based documents. Our experiments demonstrated the effectiveness of MarkupLM in handling the multi-record pages. So we set a robust base for further advancements in this area. Moreover, our research goes beyond Russian news websites and provides a valuable contribution to development of other forms of semi-structured web content extraction. 

The internet continues to expand and the ability to efficiently process multi-record web pages becomes increasingly vital too. Our research provides as a powerful tool for current applications, as a strong basis for future development.

In future work, we are going to study the extraction of additional attributes from our dataset, such as short\_text, short\_title, author, and time. These attributes, while not the focus of the current article, represent further opportunities to enhance the richness and utility of information extracted from multi-record pages. We are also going to expand our dataset with examples from other languages, for example, English.

\appendices

\ifCLASSOPTIONcaptionsoff
  \newpage
\fi

\bibliographystyle{IEEEtran}

\bibliography{main}

\section*{Additional Figures}

\end{document}